\documentclass{article}
\usepackage{ijcai26}
\usepackage{times}
\usepackage{soul}
\usepackage{url}
\usepackage[hidelinks]{hyperref}
\usepackage[utf8]{inputenc}
\usepackage[small]{caption}
\usepackage{graphicx}
\usepackage{amsmath}
\usepackage{amsthm}
\usepackage{booktabs}
\usepackage{algorithm}
\usepackage{algorithmic}
\usepackage[switch]{lineno}
\usepackage{multicol}
\usepackage{multirow}
\usepackage{tabularx}
\usepackage{float}

\title{Learning Multi-Indicator Weights for Data Selection: A Joint Task-Model Adaptation Framework with Efficient Proxies}
\author{
    Author Name
    \affiliations
    Affiliation
    \emails
    email@example.com
}
\author{
Jingze Song$^*$
\and
Zihao Chen$^*$\and
Wenqing Chen$^\dagger$\and
Zibin Zheng\\
\affiliations
School of Software Engineering, Sun Yat-sen University\\
\emails
\{songjz5, chenzh636\}@mail2.sysu.edu.cn,
\{chenwq95, zhzibin\}@mail.sysu.edu.cn
}
\begin{document}
\maketitle
{\let\thefootnote\relax\footnotetext{$^*$Equal contribution. $^\dagger$Corresponding author.}}
\begin{abstract}
    Data selection is a key component of efficient instruction tuning for large language models, as recent work has shown that data quality often matters more than data quantity. Accordingly, prior studies have introduced various multi-dimensional heuristics to evaluate and filter instruction data. However,
    most existing methods rely on static task-agnostic and model-agnostic weighting schemes, which overlook the varying requirements of specific downstream tasks and the differing pre-existing capabilities of models. In this paper, we propose a framework for learning multi-indicator weights that jointly adapts data selection to both the downstream task and the specific model. Our method identifies optimal weight configurations without full-scale fine-tuning by utilizing in-context learning (ICL) signals on compact tiny-validation sets. These signals serve as efficient performance proxies that ensure high-fidelity evaluation at minimal computational cost. Experiments across multiple benchmarks and model families, including Mistral, Qwen, and Llama, show that the approach achieves performance comparable to or exceeding full-dataset tuning while using only 30\% of the training samples on GSM8K. Furthermore, our analysis reveals a trade-off between semantic diversity and logical complexity in reasoning tasks, highlighting the necessity of joint task-model adaptation.
\end{abstract}
\section{Introduction}
Instruction tuning adjusts large language models (LLMs) to match human intent, and the choice of training data significantly influences model performance. Recent studies indicate that larger datasets may not necessarily lead to better performance, as noisy data can incur high computational costs and introduce hallucinations~\cite{chen2024automated,si-etal-2025-aligning}. Instead, the effectiveness of instruction fine-tuning mainly depends on the quality and diversity of the samples, rather than the size of the dataset.~\cite{zhou2023lima,chen2024alpagasus}.
Previous work generally evaluates instruction tuning data from three dimensions: quality, which reflects the correctness and utility of responses \cite{liu2024what,chen2024alpagasus}; complexity, which captures the logical depth or difficulty of queries \cite{li2024quantity,zhao2024tree}; and diversity, which ensures broad coverage across various domains \cite{ge2024clustering,luinstag}. Consequently, most selection pipelines aggregate these indicators using static weighting schemes or predefined manual rules~\cite{cao2024instruction}.
However, two challenges persist in these approaches. First, static weighting schemes encounter the problem of poor generalizability~\cite{diddee2025chasing} and do not account for the varying requirements of different downstream tasks~\cite{cao2024instruction} and different LLMs. The optimal weighting among indicators is universal across different tasks; for instance, mathematical reasoning tasks demand high logical complexity data, whereas general-purpose alignment may prioritize linguistic diversity. Second, while some data-driven approaches optimize the selection by identifying influential samples \cite{xia2024less,tang2025middo}, they often incur high computational costs. These methods typically rely on ``select-finetune-evaluate'' loops or high-dimensional gradient calculations to score the data. To reduce training costs, some recent approaches employ in-context learning (ICL) as a proxy for fine-tuning performance~\cite{wang-etal-2024-data}. However, they remain constrained by their sample-level granularity, suffering from high inference time costs due to the traversal of the dataset.
In this paper, we propose a joint \textbf{T}ask-\textbf{M}odel \textbf{A}daptation framework with efficient \textbf{P}roxies (TMAP) for instruction data selection. To deal with the first challenge, unlike approaches that rely on static heuristic rules, TMAP treats data selection as an optimization problem by dynamically adjusting the weighting of various indicators, to match the requirements of both the downstream task and the base model. To deal with the second challenge, TMAP introduces an efficient performance proxy that utilizes ICL signals on a tiny validation sets, bypassing the prohibitive cost of fine-tuning an LLM. By leveraging these high-fidelity proxies, TMAP can rapidly evaluate weight configurations and update them without training the model. This approach significantly reduces execution time while ensuring the selected data remains effective for target tasks.
Our main contributions are summarized as follows:
\begin{itemize}
    \item We propose a joint task-model adaptation framework that tickle the problem of static heuristics by using learned weights over multiple selection indicators to tailor data subset to specific task requirements and model capabilities.
    \item We introduce efficient proxies of select-finetune-evaluate loops via performing ICL on tiny validation sets. By utilizing validation results as proxies, we enable efficient weight-level optimization without the need for full-scale finetuning and full-validation inference.
    \item Experiments across various model families and benchmarks demonstrate that our method achieves performance comparable to full-dataset tuning, reaching considerable results on reasoning tasks such as GSM8K while utilizing only $30\%$ of the training data.
\end{itemize}
\section{Related Work}
\paragraph{Instruction data selection with different indicators}
Current indicators for instruction tuning can be broadly categorized into three types: \textit{heuristic-based}, \textit{LLM-based}, and \textit{ICL-based}. \textbf{Heuristic-based} approaches assess data quality using predefined heuristic metrics. A simple yet effective metric is selecting data with the longest output length~\cite{zhao2024long}. \cite{li2024quantity,li2024superfiltering} propose an instruction following difficulty (IFD) metric that leverages GPT-2 to compare the perplexity of model responses with and without instructions.  \textbf{LLM-based} methods employ LLMs as scorers to identify high-quality data samples. For instance, ALPAGASUS~\cite{chen2024alpagasus} and DEITA~\cite{liu2024what} use template prompts to allow LLMs to rate data for both quality and complexity.
Some LLM-based selection use \textbf{data-driven optimization} to identify influential data best aligned with model requirements. LESS \cite{xia2024less} utilizes gradients from LLM to select data, while iterative methods like Middo~\cite{tang2025middo} continually update the training LLM with selected high complexity and quality data in each iteration, aiming to select data that is aligned with the current model's needs. Additionally, LASER \cite{mirza2025laser} trains and employs data-specialized scorers to assess data quality.
\textbf{ICL-based} methods assess the quality of each candidate data point by measuring its usefulness as an ICL example for a given evaluation set. Nuggets~\cite{li2024one} computes a quality score by comparing the correctness gap on a held-out set between a zero-shot setting and a one-shot setting that uses the candidate data as the prompt. \cite{zhang2025holdout} propose an in-context approximation (ICA) approach, which estimates the loss of candidate examples by conditioning on a carefully selected in-context held-out set, and then re-weights gradient updates accordingly. Data Whisperer~\cite{wang-etal-2025-data-whisperer} randomly samples data from the candidate pool as demonstrations and queries, then computes the weighted evaluation score for each query based on the attention mechanism. The scores quantify the usefulness of the candidate data.
These data selection methods, however,  often overlook the specific needs of diverse downstream applications. Furthermore, ICL-based methods incur high computational costs for new data, and the reliability of using pre-defined or randomly sampled evaluation sets is limited. Iterative approaches typically rely on ``select-finetune-evaluate'' loops that are computationally expensive and difficult to scale.
\paragraph{Benchmark compression for efficient evaluation}
The increasing computational costs required for evaluating LLMs have spurred significant research into efficient benchmarking methods. Many studies have identified redundancy in benchmark items~\cite{ye2023predictable,perlitz2024efficient}. Consequently, methods now employ carefully selected small evaluation subsets to predict full benchmark performance. For instance, \cite{vivek2024anchor} proposes the Anchor Points method and identifies critical data points to assess models with far fewer examples \cite{polo2024tinybenchmarks}.introduces TinyBenchmarks, demonstrating that by training an Item Response Theory (IRT) model~\cite{cai2016item} on just 319 curated items can reliably estimate full performance on MMLU within 2\% error \cite{wang2025rethinking}. proposes the EssenceBench framework, which formulates benchmark compression as an optimization problem and addresses it through a combination of redundancy-based coarse filtering and an iterative genetic algorithm. Our work differs by focusing on learning task-model-adaptive weights over a rich set of indicators, while sharing only  the goal of reducing the cost of efficient evaluation.
\section{Methodology}
\subsection{Problem Formulation}
We formulate the problem of instruction data selection as follows. Given a candidate pool $\mathcal{D} = \{d_1, d_2, \dots, d_N\}$ containing $N$ samples, for each sample $d_i$, we pre-calculate $K$ heuristic-based and LLM-based indicators $\{s_{i,j}\}_{j=1}^K$. These indicators are then min-max normalized to form a feature vector $s_i \in [0, 1]^K$. Our objective is to determine an optimal weight configuration $\mathbf{w} = [w_1, w_2, \dots, w_K]^\top, w_i \in [0,1]$ to calculate a weighted score $S_i(\mathbf{w})$ for each sample:$$S_i(\mathbf{w}) = \sum_{j=1}^{K} w_j \cdot s_{i,j}$$Based on these scores, a training subset $\mathcal{D}_{\text{sub}}(\mathbf{w})$ is formed by selecting the top-$M$ samples where $M < N$. Our goal is to find an optimal $\mathbf{w}$ that maximizes the target task performance of the target LLM after instruction tuning.
\paragraph{Efficient Proxies}
Ideally, data-driven optimization for the selection problem is formulated as a select-finetune-evaluate loop. The optimization objective in this paradigm is typically denoted as follows:
\begin{equation}
    \mathbf{w^*} = \arg \max_{\mathbf{w}} \underbrace{\text{Eval} ( \underbrace{\text{SFT} (\mathcal{M}, \mathcal{D}_{\text{sub}}(\mathbf{w}))}_{\text{High-cost}}, \mathcal{V}_{\text{full}} )}_{\text{High-cost}}
\end{equation}
where $\text{SFT}(\cdot)$ denotes the supervised fine-tuning of a LLM $\mathcal{M}$ on the selected subset, and $\text{Eval}(\cdot, \mathcal{V}_{\text{full}})$ represents evaluating on the full downstream validation set. However, this process is computationally prohibitive due to the repeated LLM training and large-scale evaluation inference. To address these challenges, we redefine the selection process using two efficient proxies: ICL and tinybenchmark. Instead of performing full-scale fine-tuning, we utilize few-shot ICL to measure the utility of $\mathcal{D}_{\text{sub}}(\mathbf{w})$. Instead of the full validation set $\mathcal{V}_{\text{full}}$, we evaluate performance on a tiny validation set $\mathcal{V}_{\text{tiny}} \subset \mathcal{V}_{\text{full}}$. We define a reward function $R(\mathbf{w})$ that serves as a proxy for the fine-tuning outcome:
\begin{equation}
R(\mathbf{w}) = \text{Eval}\left(\text{ICL}(\mathcal{M}, \mathcal{D}_{\text{sub}}(\mathbf{w})), \mathcal{V}_{\text{tiny}}\right)
\end{equation}
By shifting the optimization target to $\arg \max_{\mathbf{w}}  R(\mathbf{w})$, the framework finds approximately optimal configurations at the indicator level while reducing computational costs and execution time. The remaining problem is how to guarantee that the performance approximation via efficient proxies is reasonable. We introduce the details in Section \ref{sec:proxies}
\begin{figure*}[t]
\centering
\includegraphics[width=\textwidth]{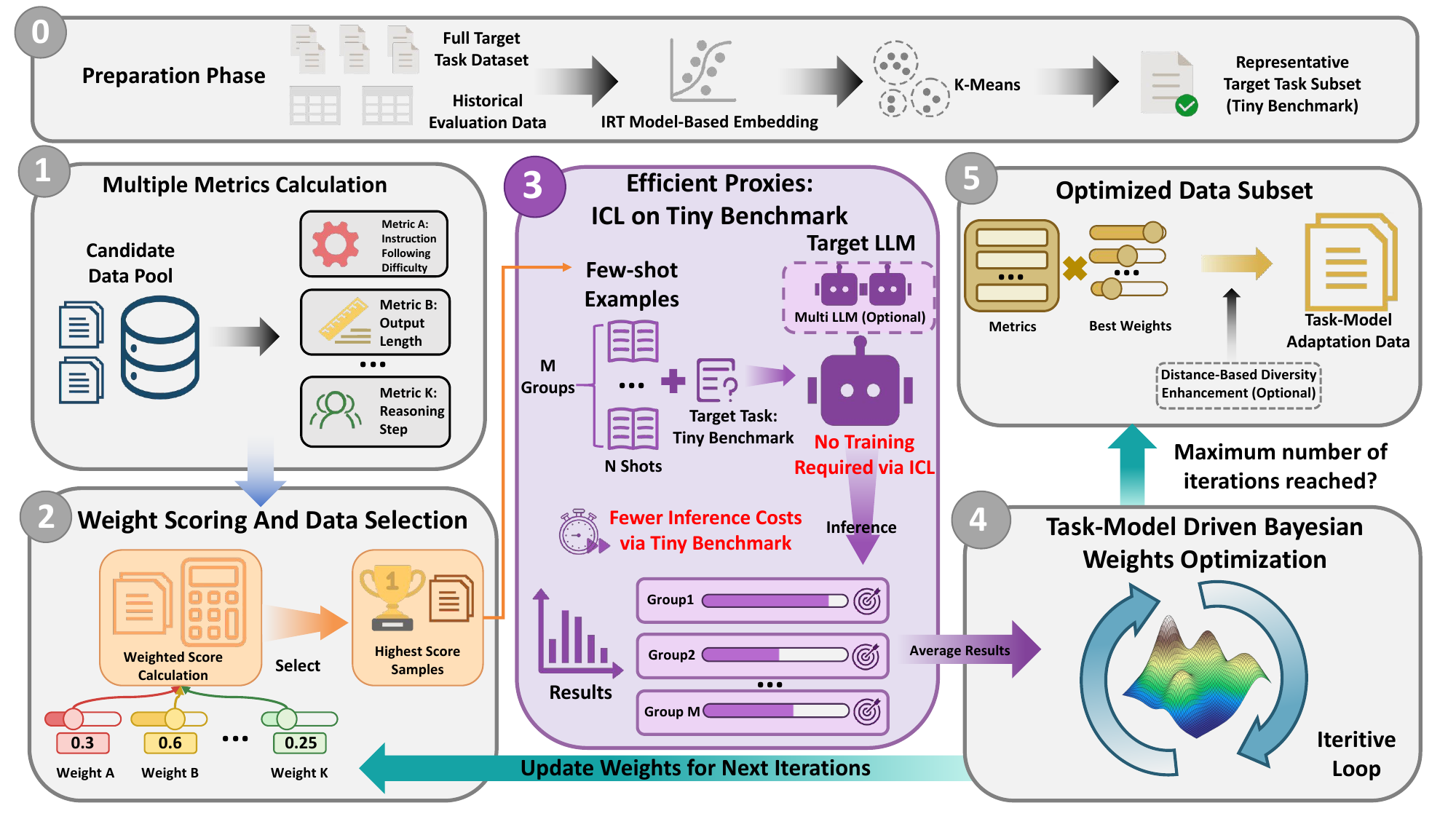}
\caption{Overview of the Task-Model Adaptation with efficient Proxies (TMAP) data selection pipeline. The framework scores candidate instructions along multiple dimensions, optimizes task-specific weights, and selects a high-value subset for downstream fine-tuning.}
\label{fig:TMAP_overview}
\end{figure*}
\subsection{Overview of TMAP}
As show in Figure \ref{fig:TMAP_overview}, we treat data selection as a weight optimization problem. The TAMP pipeline consists of: (1) sampling a tiny validation set via K-means on IRT embeddings; (2) scoring the candidate data pool across an extensible set of indicators; (3) calculating weighted scores and selecting the highest-scoring data to form an ICL demonstration set; (4) iteratively optimizing indicator weights using Bayesian Optimization, guided by an ICL-based reward signal on the tiny validation set; and (5) selecting the final subset based on the optimized weights once optimization finished.
\subsection{Multi-dimensional Indicators}
To capture multiple features of data quality, we score each candidate sample $d_i$ with $K$ dimensions to form a normalized feature vector $s_i \in [0, 1]^K$. Following by prior work, we implement four key indicator groups: (1) \textbf{Semantic Quality}: LLM-based scores \texttt{alpagasus\_score} and \texttt{deita\_quality\_score}, which capture the faithfulness and helpfulness of responses~\cite{chen2024alpagasus,liu2024what}. (2) \textbf{Complexity and Difficulty}: \texttt{ifd} and \texttt{deita\_complexity\_score}, which measure how challenging an instruction-response pair is for the model~\cite{li2024superfiltering,liu2024what}. (3) \textbf{Linguistic Heuristics}: \texttt{output\_length}~\cite{zhao2024long} metric, which aligns with general-purpose filtering. (4) \textbf{Domain-specific metrics}: For the GSM8K dataset, we use \texttt{reason\_steps}~\cite{jin-etal-2024-impact} to approximate the length of the underlying chain-of-thought. Note that TMAP is easy to adapt to different domains (e.g., dropping reasoning metrics for dialogue tasks).
\subsection{Efficient Proxies}
\label{sec:proxies}
\paragraph{Proxy for Finetuning}
To avoid the high computation cost of repeated SFT in the optimization cycle, we leverage ICL as an efficient proxy. Recent studies confirm that a sample's utility as a few-shot demonstration strongly correlates with its utility as a training instance~\cite{yin2024deeper,zhaocontext}. Note that, unlike prior work, we do not replicate exact SFT performance. Instead, our goal is to preserve the relative ranking of data quality. As shown in Appendix A, varying the number of ICL demonstrations consistently maintain the same performance trends as observed in SFT training. This demonstrates that ICL can serve as a reliable proxy of SFT in evaluating the influence of different data on the target model. To reduce the impact of ICL prompt order sensitivity on the reward signal $R(\mathbf{w})$, we partition the selected subset $\mathcal{D}_{\text{sub}}$ into $m$ distinct groups of $n$-shot exemplars. We then evaluate the base model using these groups as prompts, deriving the reward from the aggregated accuracy.
\paragraph{Proxy for Full-validation Evaluating}
To avoid time-consuming full-set inference, we construct a compact \textbf{tinybenchmark} $\mathcal{V}_{\text{tiny}}$ of $S$ representative samples. Our target is to approximate the full-set performance via weighted evaluation on this subset:
\begin{equation}
\small
\label{eq:acc}
  R(\mathbf{w})=\frac{1}{|\mathcal{V}_{\text{full}}|}\sum_{v\in \mathcal{V}_{\text{full}}}Y_{v} \approx \sum_{v\in \mathcal{V}_{\text{tiny}}} w_vY_{v}  .
\end{equation}
$Y_{v\mathcal{M}}\in \{0,1\}$ is the correctness of sample $v$ and $w_v$ is the weight of sample $v$. The underlying principle is that an LLM can accurately answer a broad range of questions if and only if it correctly answers these representative samples.
To select representative samples, we employ IRT to generate embedding vectors.  In the context of TMAP, we adopt the Multidimensional Two-Parameter Logistic (M-2PL) model from IRT. The M-2PL model assumes that the probability of the LLM $\mathcal{M}$ correctly solve the validation sample $v$ is:
\begin{equation}
\small
\label{eq:m2pl}
    p_{vl}=\frac{1}{1+exp(-(\alpha_v^\top \theta_{\mathcal{M}}+\beta _v  ))},
\end{equation}
where $\theta_\mathcal{M}$ represents latent LLM abilities and $(\alpha_v, \beta_v)$ represent item difficulty and discrimination. We concatenate the point estimates parameters to form embedding vectors $E_v=(\hat{\alpha}, \hat{\beta})$ for sample $v$. We apply K-means clustering to identify centroids and select the item $v^*$ closest to form $V_{tiny}$:
\begin{equation}
\mathcal{V}_{\text{tiny}} = \{v^*_1, \dots, v^*_o\}
\end{equation}
To further refine precision of tinybenchmark, following~\cite{polo2024tinybenchmarks}, we combine the observed results and IRT prediction:
\begin{equation}
\small
\begin{aligned}[t]
    R(\mathbf{w}) & = \lambda\sum_{i \in \mathcal{V}_{\text{tiny}}}w_i Y_{i}+
     (1-\lambda)  (\delta_1 \sum_{i \in \mathcal{V}_{\text{tiny}}} Y_{i} + \delta_2 \sum_{j \notin \mathcal{V}_{\text{tiny}}} \hat{p}_{i} )
\end{aligned}
\end{equation}
where $\lambda$ is a learnable parameters, $\delta_1= \frac{\hat{\lambda}}{|\mathcal{V}_{\text{tiny}}|}$, $\delta_2=\frac{1-\hat{\lambda}}{|\mathcal{V}_{\text{full}} \setminus \mathcal{V}_{\text{tiny}}|}$ and $\hat{\lambda}=|\mathcal{V}_{\text{tiny}}|/| \mathcal{V}_{\text{full}}|\in [0,1]$.
Experiment in Appendix B proves the reliability of the proxy for full-validation evaluating.
\subsection{Task-Model Adaptive Weight Optimization}
We formulate data selection as a black-box optimization problem: $\mathbf{w}^* = \arg\max_{\mathbf{w}} R(\mathbf{w})$, where $R(\mathbf{w})$ represents the proxy reward signal. To efficiently solve this non-differentiable objective, we employ Bayesian Optimization (BO)~\cite{movckus1974bayesian}.
As outlined in Algorithm~\ref{alg:TMAP}, the process iteratively updates weights $\mathbf{w}$. In each iteration $t$, we compute weighted scores $S_i = \sum_{j=1}^K w_{t,j} \cdot s_{i,j}$ for all candidates, select the top-$M$ instances, and partition them into $m$ groups. These groups are evaluated via ICL on $\mathcal{V}_{\text{tiny}}$ to obtain a robust reward signal $R_t$, which guides the Bayesian optimizer to propose the next configuration $\mathbf{w}_{t+1}$.
\begin{algorithm}[tb]
\caption{Task-Adaptive Weight Optimization}
\label{alg:TMAP}
\begin{algorithmic}[1]
\STATE {\bfseries Input:} Candidate pool $\mathcal{D}$, Indicators $\{s_j\}_{j=1}^K$, Target task tiny validation set $\mathcal{V}_{\text{tiny}}$, Selection size $M$
\STATE {\bfseries Initialize:} Weights $\mathbf{w}_0$, Bayesian Optimizer $\mathcal{B}$
\FOR{$t = 1$ \textbf{to} $T$}
    \STATE $S_i \leftarrow \sum_{j=1}^K w_{t,j} \cdot s_{i,j}$ for each $d_i \in \mathcal{D}$
    \STATE $\mathcal{D}_{sub} \leftarrow \text{Top-M}(\mathcal{D}, S_i)$
    \STATE $\{\mathcal{D}_{\text{sub}_l}\}_{l=1}^m \leftarrow \text{Partition}(\mathcal{D}_{\text{sub}}) $
    \STATE $R_t \leftarrow \text{EvalICL}(\{\mathcal{D}_{\text{sub}_l}\}_{l=1}^m, \mathcal{V}_{\text{tiny}})$ \COMMENT{Average performance over $m$ groups}
    \STATE $\mathbf{w}_{t+1} \leftarrow \text{Update}(\mathbf{w}_t, R_t)$
\ENDFOR
\STATE {\bfseries Return} $\mathbf{w}^*$
\end{algorithmic}
\end{algorithm}
\subsection{Final Filtering and Robustness Refinement }
To capture the multifaceted nature of data quality, we score each candidate sample $d_i$ across $K$ dimensions to form a normalized feature vector $\mathbf{s}_i \in [0, 1]^K$. Drawing from prior work, we implement four key indicator groups:
\begin{itemize}
    \item \textbf{Semantic Quality}: LLM-based scores, specifically \texttt{alpagasus\_score}~\cite{chen2024alpagasus} and \texttt{deita\_quality\_score}~\cite{liu2024what}, which capture the faithfulness and helpfulness of responses.
    \item \textbf{Complexity and Difficulty}: Metrics like \texttt{ifd} and \texttt{deita\_complexity\_score}, which measure how challenging an instruction-response pair is for the model~\cite{li2024superfiltering,liu2024what}.
    \item \textbf{Linguistic Heuristics}: Simple statistics such as \texttt{output\_length}~\cite{zhao2024long}, consistent with general-purpose filtering metrics.
    \item \textbf{Domain-Specific Metrics}: For reasoning tasks like GSM8K, we include \texttt{reason\_steps}~\cite{jin-etal-2024-impact} to approximate the length of the underlying chain-of-thought.
\end{itemize}
Note that TMAP is designed to be extensible, allowing for easy adaptation to different domains (e.g., excluding reasoning metrics for dialogue tasks).
\begin{table*}[t]
\centering
\resizebox{\linewidth}{!}{
\begin{tabular}{@{}lccclccclccc@{}}
\toprule
 & \multicolumn{3}{c}{\textbf{Mistral-v0.2-7B}} & \multicolumn{3}{c}{\textbf{Qwen2.5-7B}} & \multicolumn{3}{c}{\textbf{Llama3-8B}} \\
\cmidrule(lr){2-4} \cmidrule(lr){5-7}  \cmidrule(lr){8-10}
\textbf{Method} & \textbf{GSM8K} & \textbf{MUSR} & \textbf{GPQA}  & \textbf{GSM8K} & \textbf{MUSR} & \textbf{GPQA} & \textbf{GSM8K} & \textbf{MUSR} & \textbf{GPQA} \\
\midrule
Base & 38.67 & 47.49 & \textbf{28.86} & \textbf{79.08} & 43.67 & 32.39 & \textbf{72.10} & 37.96 & 31.29 \\
Full (100\%) & \textbf{58.61} & 46.30 & 28.78  & 75.36 & \textbf{46.03} & \textbf{32.89}  & 64.67 & 40.21 & \textbf{31.46} \\
Random & 50.80 & 46.16 & 27.68 & 75.28 & 43.77 & 31.55 & 61.49 & 41.67 & 30.12    \\
\midrule
\textit{Metric-based Methods} \\
IFD & 52.08 & 44.44 & 27.52 & 77.33 & 44.84 & 32.38 & 60.02 & 42.72 & 30.17 \\
Longest & 51.25 & 45.90 & 27.68 & 78.70 & 43.92 & 31.71 & 60.65 & 37.30 & 30.03 \\
\midrule
\textit{LLM-based Methods} \\
Alpagasus & 52.77 & 46.43 & 27.36 & 76.57 & 44.02 & 32.21 & 60.88 & 42.86 & 30.08 \\
DEITA & 52.01 & 45.44 & 27.18 & 76.27 & 43.52 & 31.88 & 62.70 & 38.89 & 30.63 \\
\midrule
\textit{ICL-based Methods} \\
Data-Whisperer & 49.58 &  47.89 & 27.77 & 78.77 & 44.56 &  32.55 & 60.73 & 41.80  & 30.20 \\
\midrule
\textbf{TMAP (Ours)} & \underline{55.42} & \textbf{\underline{50.77}} & \underline{27.94} & \underline{79.00} & \underline{45.90} & \underline{32.74} & \underline{65.96} & \textbf{\underline{42.99}} & \underline{30.73}  \\
\bottomrule
\end{tabular}
}
\caption{Performance on GSM8K/MUSR/GPQA datasets across different models. The best overall score in each column is highlighted in \textbf{bold}, while the best score among data selection methods (excluding Base and Full) is \underline{underlined}. We omit the ``Instruct'' suffix in the model names for brevity. Base refers to the original model. Full means finetuning the model using the complete training set.}
\label{tab:exp1}
\end{table*}
\section{Experiments}
\subsection{Experimental Setup}
\paragraph{Datasets}
We focus on knowledge and reasoning benchmarks that are widely used for evaluating instruction-tuned LLMs. Specifically, we used the following datasets. \textbf{GSM8K}, a dataset of grade-school math word problems requiring multi-step reasoning~\cite{cobbe2021training}. \textbf{ARC-C}, a challenging multiple-choice science QA subset with strong distractors~\cite{clark2018think}. \textbf{MUSR}, a dataset for evaluating LLMs on multistep soft reasoning tasks specified in a natural language narrative~\cite{spraguemusr}. \textbf{GPQA}, a multiple-choice dataset of hard questions created by experts in biology, physics, and chemistry. For the synthetic-data experiments, we additionally construct DeepSeek-distilled variants of GSM8K and ARC-C by augmenting each example with model-generated chain-of-thought solutions, following the setup summarized in Section~\ref{tab:exp2}.
\subsection{Baselines}
We compared TMAP with several widely used data selection techniques: (i) Random, which randomly samples subsets from the dataset. (ii) IFD, which uses GPT-2-computed IFD scores to select high-quality samples. (iii) Longest, which selects data with longest output length. (iv) Alpagasus, which prompts a LLM with a hand-craft template to score and select data. (v) DEITA, which selects data with highest quality, complexity, and diversity. (vi) Data Whisperer, which leverages ICL and attention scores to select data.
\subsection{Implementation Details}
We use four backbone models of different scales and families: Mistral-v0.2-7B-Instruct~\cite{jiang2023mistral}, Qwen2.5-7B-Instruct, Llama3-8B-Instruct and Llama2-13B-Chat. For GSM8K and ARC-C, we train the model using 30\% samples from their corresponding training set, aligning with~\cite{wang-etal-2025-data-whisperer}. For GPQA and MUSR, we use 10\% samples from DS2-50K dataset for training following the settings of prior work~\cite{pangtoken}. We report the Exact Match (EM) accuracy on the test set for GSM8K and the other three tasks are reported with their multi-choice accuracy. All experiments were performed on 8 NVIDIA A100 GPUs. We utilized LoRA~\cite{hu2022lora} for model fine-tuning. During the Bayesian optimization process, we set the ICL evaluation parameters to $m=5$ groups and $n=3$ exemplars per group. We employ 5 random initialization steps followed by 200 optimization iterations. Further details can be found in supporting materials.
\subsection{Main Results: Effectiveness of TMAP}
We evaluate the effectiveness of TMAP through multiple settings, including cross-model validation (Exp 1), synthetic data distillation (Exp 2), and multi-model generalization (Exp 3).
\subsubsection{Performance Across Models and Tasks}
As shown in Table~\ref{tab:exp1}, TMAP consistently outperforms heuristic baselines across all models. Notably, on Qwen2.5-7B with GSM8K, it achieves comparable performance to baselines while using only 30\% of training data.
These results demonstrate that learning task-specific weights for multi-dimensional heuristics is more effective than relying on a single fixed metric or uniform weighting.
\subsubsection{Validation on Synthetic Data}
In Exp 2, we evaluate TMAP on distilled datasets where reasoning paths are augmented using \textbf{DeepSeek}~\cite{liu2025deepseek}.
We expand our evaluation to include both the mathematical reasoning benchmark \textbf{GSM8K} and the science reasoning benchmark \textbf{ARC-C}.
For this setting, we select 30\% of the synthetic data pool for fine-tuning Llama2-13B and Mistral-v0.2-7B.
As shown in Table~\ref{tab:exp2}, TMAP consistently outperforms other data selection baselines.
For Llama2-13B on GSM8K, TMAP achieves a score of \textbf{52.77}, surpassing DEITA (51.78) and Alpagasus (49.28).
More notably, on the ARC-C benchmark using Mistral-v0.2-7B, our method reaches \textbf{77.05}, which is virtually equivalent to the performance of fine-tuning on the full synthetic dataset (77.22), despite using only 30\% of the training samples.
This result highlights TMAP's capability to filter high-quality reasoning traces from DeepSeek-generated data, effectively identifying the ``gold'' within synthetic pools while discarding redundant or lower-quality generations.
\begin{table}[t]
\centering
\resizebox{\linewidth}{!}{
\begin{tabular}{@{}lcccc@{}}
\toprule
 & \multicolumn{2}{c}{\textbf{Llama2-13B}} & \multicolumn{2}{c}{\textbf{Mistral-v0.2-7B}} \\ \cmidrule(lr){2-3} \cmidrule(lr){4-5}
\textbf{Method} & \textbf{GSM8K} & \textbf{ARC-C} & \textbf{GSM8K} & \textbf{ARC-C} \\ \midrule
Base & 35.48 & 60.58 & 38.59 & 56.65 \\
Full Synthetic (100\%) & \textbf{59.51} & \textbf{67.23} & \textbf{75.36} & \textbf{77.22} \\ \midrule
Random & 48.52 & 64.33 & 67.48 & 69.37 \\
 IFD & 48.67 & 64.76 & 71.04 &74.15 \\
Longest& 49.66& 64.25& 70.89& 74.40\\
Alpagasus & 49.28 & 64.51 & 67.40 & 76.45 \\
DEITA & 51.78 & 65.52 & 70.89 & 75.10 \\
\textbf{TMAP (Ours)} & \underline{52.77} & \underline{66.13} & \underline{71.34} & \underline{77.05} \\ \bottomrule
\end{tabular}
}
\caption{Filtering performance on DeepSeek-distilled GSM8K and ARC-C datasets (30\% selection). The best overall score in each column is highlighted in \textbf{bold}, while the best score among data selection methods (excluding Base and Full) is \underline{underlined}. TMAP shows improvements over the heuristic baselines across both reasoning tasks in our experiments.}
\label{tab:exp2}
\end{table}
\subsubsection{Multi-Model Generalization}
In Exp 3, we replace the single-model reward with a multi-model consensus signal obtained by averaging the ICL-based rewards from Mistral-v0.2-7B, Qwen2.5-7B, and Llama3-8B. As shown in Table~\ref{tab:exp3_multi_model}, this consensus reward leads to slightly lower peak scores on each individual model compared to the single-model optimal configuration.
However, the consensus weights remain competitive with strong baselines and, more importantly, capture a more robust data mixture that transfers well across architectures. This highlights the cross-model generalizability of TMAP: even when no single model is explicitly favored during optimization, the learned weights deliver stable gains across Mistral, Qwen2.5, and Llama3.
\begin{table}[h]
\centering
\resizebox{\linewidth}{!}{
\begin{tabular}{@{}lccc@{}}
\toprule
\textbf{Reward Scheme} & \textbf{Mistral-v0.2-7B} & \textbf{Qwen2.5-7B} & \textbf{Llama3-8B} \\ \midrule
Single-Model Opt. (S)  & \textbf{55.42}      & \textbf{79.00}      & \textbf{65.96}     \\
Multi-Model Opt. (M)   & 53.60               & 78.46               & 64.06              \\ \midrule
\textit{Delta (M - S)} & \textit{-1.82}      & \textit{-0.54}      & \textit{-1.90}     \\ \bottomrule
\end{tabular}
}
\caption{Comparison between Single-Model (S) reward and Multi-Model (M) consensus reward. While single-model optimization yields the highest scores on each model, multi-model rewards provide more stable cross-model behavior in our experiments.}
\label{tab:exp3_multi_model}
\end{table}
\subsection{Ablation Study: Impact of Diversity}
In Exp 4, we investigate the impact of incorporating diversity-based filtering (following the DEITA approach \cite{liu2024what}) after determining the optimal weights via TMAP.
Contrary to the prevailing intuition that diversity is always beneficial, our results in Table~\ref{tab:exp4_diversity} reveal a consistent performance decline across all tested models when explicit diversity constraints are enforced. Specifically, the accuracy for Mistral drops by 4.17\%, while Llama3 and Qwen2.5 see reductions of 1.90\% and 0.38\%, respectively. This confirms our hypothesis that for reasoning-dense tasks, the priority should be logic-complexity rather than semantic surface variation. In such reasoning-dense settings, explicit diversity filtering may prune semantically similar yet logically distinct hard examples that are crucial for strengthening step-by-step mathematical reasoning.
\begin{table}[h]
\centering
{
\begin{tabular}{@{}lccc@{}}
\toprule
\textbf{Model} & \textbf{TMAP (BO)} & \textbf{TMAP + Diversity} & \textbf{Delta} \\ \midrule
Mistral-v0.2-7B     & \textbf{55.42}     & 51.25                     & -4.17          \\
Qwen2.5-7B     & \textbf{79.00}     & 78.62                     & -0.38          \\
Llama3-8B      & \textbf{65.96}     & 64.06                     & -1.90          \\ \bottomrule
\end{tabular}
}
\caption{Ablation study on diversity-enhanced selection across different models. Incorporating explicit diversity filtering (similarity-threshold 0.9) consistently leads to a performance drop in reasoning tasks.}
\label{tab:exp4_diversity}
\end{table}
\subsection{Ablation Study: Optimal Granularity $m$}
To investigate the impact of partitioning granularity on reward signal stability, we evaluate our method across various $m \in \{1, 2,4, 8, 16\}$ on the MUSR datasets using Qwen2.5-7B.
\begin{table}[h]
\centering
\begin{tabular}{@{}ccc@{}}
\toprule
\textbf{Granularity ($m$)} & \textbf{MUSR}    \\ \midrule
$16$  & 45.25  \\
$8$  & 46.00    \\
$4$  & 44.50   \\
$2$   & 44.50  \\
$1$   & 44.05  \\\bottomrule
\end{tabular}
\caption{Performance comparison across different granularities $m$ on GPQA and MUSR. Results are accuracy (\%). }
\label{tab:granularity_results}
\end{table}
\paragraph{Analysis of Granularity $m$.}
As illustrated in Table~\ref{tab:granularity_results}, the choice of granularity $m$ significantly influences reasoning performance, exhibiting an ``inverted U-shape'' trend. We observe that performance improves as $m$ increases from 1 to 8, peaking at 46.00\%, before declining at $m=16$. This suggests that extreme granularities are suboptimal: overly coarse granularity (e.g., $m=1$) fails to capture sufficient variance for robust ranking, while excessively fine granularity (e.g., $m=16$) introduces noise that masks underlying weight patterns, thereby hindering precise optimization. Considering the trade-off between performance and computational efficiency, we adopt $m=5$ in our main experiments. Notably, the superior performance observed at $m=8$ implies that our reported results do not represent the absolute ceiling, indicating that TMAP possesses an even higher performance upper bound.
\subsection{Efficiency Analysis}
To assess the computational scalability of TMAP, we benchmark its overhead against three baselines: (1) \textbf{Full-set Evaluation}, utilizing the complete validation set for ICL reward calculation; (2) \textbf{Data Whisperer}, which computes scores by traversing the entire candidate dataset;  (3) \textbf{Standard SFT}, representing the wall-clock time for a single fine-tuning epoch. To simulate resource-constrained environments, all benchmarks are conducted on two NVIDIA A100 GPUs using the Llama3-8B backbone.
As presented in Table~\ref{tab:efficiency}, TMAP ($m=5$) requires approximately \textbf{3 minutes} per optimization iteration. A defining advantage of TMAP is its scalability regarding candidate pool size. While Data Whisperer exhibits linear complexity, where runtime surges from 11 hours to over 72 hours as the pool expands from GSM8K (7.5k) to DS2-50K (50k), TMAP maintains a nearly constant optimization cost ($\approx$ 3.0 to 3.5 minutes per iteration) across both settings.
This efficiency stems from the decoupling of the evaluation proxy from the size of the training set. TMAP's computational cost is determined solely by the inference complexity of the target task (specifically, generation length and the size of $\mathcal{V}_{\text{tiny}}$). By evaluating weight configurations via a compact proxy rather than scoring individual samples, TMAP achieves high-fidelity selection with significantly reduced overhead, facilitating scalability to massive instruction tuning datasets.
\begin{table}[t]
\centering
\small
\resizebox{\linewidth}{!}{
\begin{tabular}{@{}lcccc@{}}
\toprule
\textbf{Method} & \textbf{Granularity} & \textbf{Cost / Iter.} & \textbf{Total Cost} \\
& & & \textit{(200 Iters)} \\
\midrule
Standard SFT (1 Epoch) & - & 66 min &  220 h\\
Full-set Evaluation & $m=5$ & 38 min &  127 h\\
\midrule
Data-Whisperer (GSM8K) & - & - &  11 h\\
Data-Whisperer (DS2-50K) & - & - & $> 72$ h\\
\midrule
\textbf{TMAP (GSM8K)}& $m=5$ & \textbf{3.0 min} & \textbf{10 h}\\
\textbf{TMAP (DS2-50K)}& $m=5$ & \textbf{3.5 min} & \textbf{12 h}\\
\bottomrule
\end{tabular}
}
\caption{Efficiency comparison on Llama3-8B. Unlike sample-level scoring methods, \textbf{TMAP maintains a stable time cost regardless of candidate dataset size} (GSM8K vs. DS2-50K), as its complexity depends solely on the size of the tiny validation set $\mathcal{V}_{\text{tiny}}$.}
\label{tab:efficiency}
\end{table}
\begin{figure}[t]
\centering
\includegraphics[width=1.0\columnwidth]{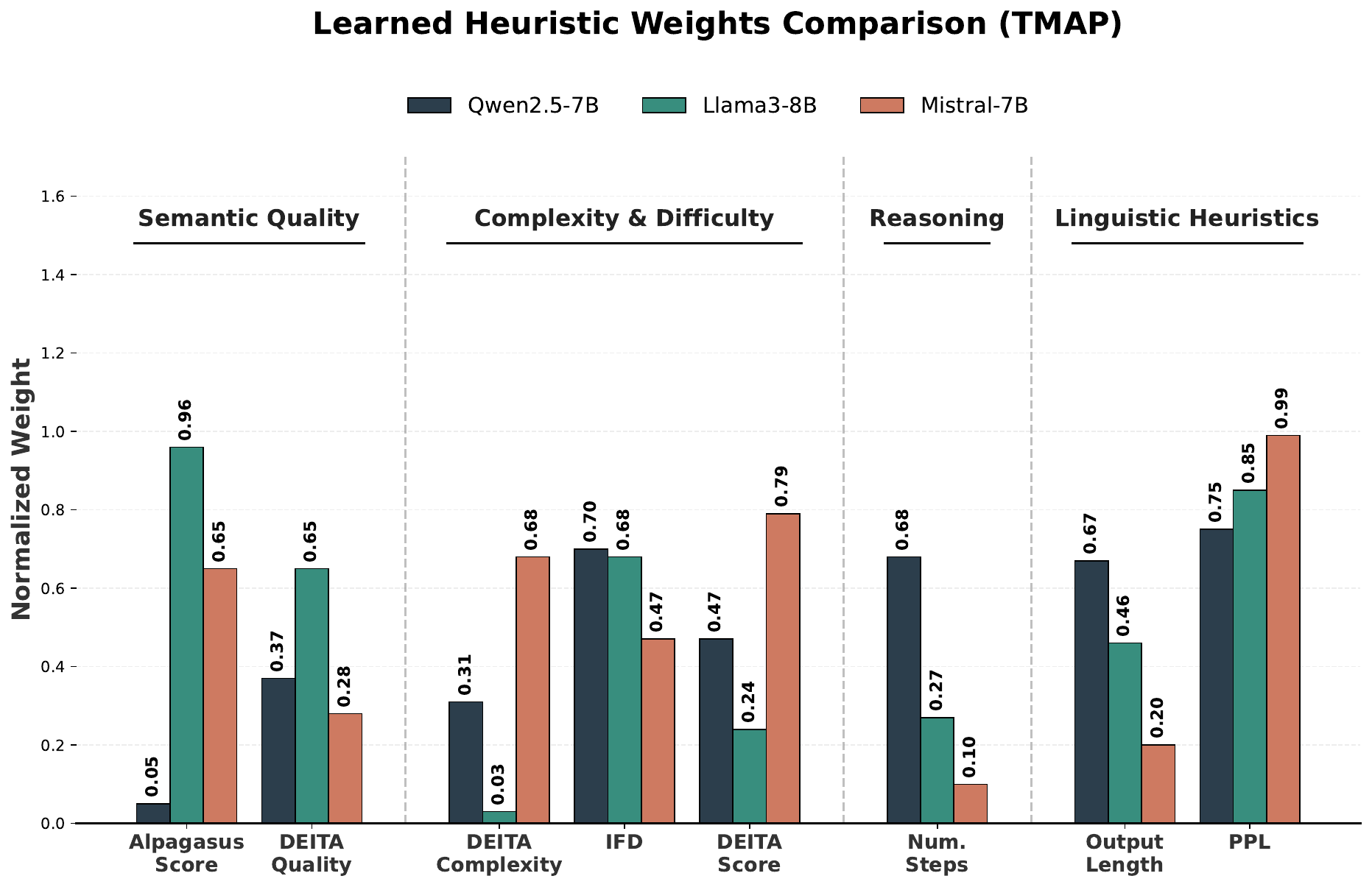}
\caption{Column chart visualization of learned heuristic weights across Qwen2.5, Llama3, and Mistral on GSM8K.}
\label{fig:learned_weights_radar}
\end{figure}
\section{Weight Interpretation and Analysis}
A core contribution of TMAP is discovering autonomous, model-specific weighting schemes. We analyze the converged weights for Qwen2.5, Llama3, and Mistral on GSM8K to understand their distinct definitions of ``high-quality'' data.
\subsection{Model-Specific Heuristic Preferences}
\begin{table}[t]
\centering
\small
\resizebox{\linewidth}{!}{
\begin{tabular}{@{}lccc@{}}
\toprule
\textbf{Indicator} & \textbf{Qwen2.5} & \textbf{Llama3} & \textbf{Mistral} \\ \midrule
Alpagasus Score \cite{chen2024alpagasus} & 0.050 & 0.965 & 0.649 \\
DEITA Complexity \cite{liu2024what}    & 0.314 & 0.029 & 0.675 \\
DEITA Quality \cite{liu2024what}       & 0.371 & 0.652 & 0.281 \\
DEITA Score (Joint) \cite{liu2024what} & 0.468 & 0.237 & 0.787 \\
IFD (PPL-based) \cite{li2024superfiltering} & 0.700 & 0.675 & 0.467 \\
Number of Reasoning Steps~\cite{jin-etal-2024-impact}                              & 0.679 & 0.271 & 0.101 \\
Output Length~\cite{zhao2024long}        & 0.674 & 0.458 & 0.199 \\
Perplexity (PPL) \cite{jin2024optimizing} & 0.752 & 0.854 & 0.991 \\ \bottomrule
\end{tabular}
}
\caption{Learned optimal weights for multi-dimensional heuristics across different base models on GSM8K. Values are normalized to a $[0, 1]$ scale.}
\label{tab:learned_weights}
\end{table}
Table~\ref{tab:learned_weights} and Figure~\ref{fig:learned_weights_radar} show that optimal weight configurations vary significantly, confirming that data quality is relative to the model's architecture rather than a static property.
\paragraph{Qwen2.5: Complexity and Length.}
Qwen2.5 assigns high weights to \textit{IFD} (0.700), \textit{Reasoning Steps} (0.679), and \textit{Output Length} (0.674), while minimizing \textit{Alpagasus Score} (0.050). As a strong reasoner, Qwen benefits most from structural complexity and hard data, relying less on external semantic judgments.
\paragraph{Llama3: Reliance on Strong Judges.}
Llama3 heavily weights \textit{Alpagasus Score} (0.965) while ignoring \textit{DEITA Complexity} (0.029). This indicates a priority for correctness and clear instruction-following over raw difficulty. TMAP steers selection toward authoritative samples to avoid introducing noise during alignment.
\paragraph{Mistral: High Information Density.}
Mistral prioritizes \textit{Perplexity} (0.991) and \textit{DEITA Score} (0.787) but ignores \textit{Reasoning Steps} (0.101). Given the high-quality candidate pool, this preference for high perplexity suggests Mistral benefits from information density or surprisal, requiring data that diverges from its pre-training priors rather than predictable patterns.
\subsection{The Necessity of Task-Adaptive Weighting}
The fact that \textit{DEITA Complexity} is prioritized by Mistral (0.675) but largely ignored by Llama3 (0.029) highlights a critical trade-off: a sample that is ``appropriately difficult'' for one model might be redundant or incorrectly scored for another. By bypassing manual rule-setting, TMAP identifies these hidden alignment points between data characteristics and model capabilities.
\section{Conclusion}
In this paper, we introduced TMAP, a joint task-model adaptation framework designed for instruction tuning data selection, which overcomes the limitations of static heuristic weighting and the high computation costs of existing data selection methods. By combining Bayesian Optimization with two efficient proxies In-Context Learning (ICL) and tinyBenchmark, TMAP iteratively searches optimal, multi-dimensional indicator weights tailored to specific target tasks and model. Experiments across various model families and benchmarks demonstrate that TMAP significantly outperforms static baselines with high computational efficiency. These results validate the necessity of adaptive, task-specific data selection strategies in instruction tuning. Future work will focus on scaling TMAP to more diverse instruction-tuning categories, such as creative writing and code generation.
\section*{Acknowledgments}
This work was supported by the National Natural Science Foundation of China (62306344), Guangdong Basic and Applied Basic Research Foundation (2024A1515010253, 2026A1515011800), and the Open Research Fund of the State Key Laboratory of Blockchain and Data Security, Zhejiang University.

\bibliographystyle{named}
\bibliography{ijcai26}

\clearpage %
\appendix
\section{Performance comparison between SFT and ICT}
\begin{table}[h]
\centering
\resizebox{\linewidth}{!}{
\begin{tabular}{@{}lcccc@{}}
\toprule
\textbf{Shots} & \textbf{GPQA ICL} & \textbf{GPQA SFT} & \textbf{MUSR ICL} & \textbf{MUSR SFT} \\ \midrule
$64$  & 32.47 & 32.51 & 42.59 & 42.72 \\
$32$  & \textbf{33.05} & \textbf{32.89} & \textbf{44.05} & \textbf{42.99} \\
$16$  & 31.88 & 32.21 & 43.52 & 42.86 \\
$8$   & 32.38 & 32.55 & 43.39 & 42.59 \\ \bottomrule
\end{tabular}
}
\caption{Performance comparison across different shots on GPQA and MUSR. Results are accuracy (\%). The best score is highlighted in bold. }
\label{tab:sft_ict}
\end{table}
We use the same quantity of data to prompt and fine-tune the Qwen2.5-7B model respectively. As shown in the table~\ref{tab:sft_ict}, the performance trends of SFT and ICL are consistent across different shot settings. This experimental result demonstrates that ICL can serve as a proxy for SFT. Data that achieves higher scores after SFT training also performs well in ICL, indicating that ICL can identify high-quality data and provide a stable reward signal.
\section{Performance comparison Tinybenchmark and Full Evaluation Benchmark}
\begin{table}[h]
\centering
\resizebox{\linewidth}{!}
{
\begin{tabular}{@{}lccccc@{}}
\toprule
\small
\textbf{Model} &  & \textbf{GPQA} & \textbf{Delta} & \textbf{MUSR} & \textbf{Delta} \\ \midrule
\multirow{3}{*}{Mistral-7B} &  Full Benchmark  & 28.86 &  -  &  47.49                   & -        \\
& Random         & 32.20     &   3.34     & 29.73              & 17.76          \\
& K-means-Only   & 34.00     &  5.14       & 49.00          & 1.51          \\
& Tinybenchmark  & 30.72   & \textbf{1.86}   & 46.36        & \textbf{1.13}         \\
 \midrule
\multirow{3}{*}{Qwen2.5-7B }  & Full Benchmark& 32.39  & -  &  43.67        & -         \\
& Random        & 33.90  & 1.51  & 32.00                    &   11.67       \\
& K-means-Only  & 36.00  &  3.61 & 46.00                     & 2.33         \\
& Tinybenchmark & 35.89  &  \textbf{3.50} & 41.53                     & \textbf{2.14}          \\
 \midrule
\multirow{3}{*}{Llama3-8B } &  Full Benchmark  & 31.29  & -   & 37.96                    & -        \\
& Random                   & 37.29             & 6.00          & 29.73   & 8.23      \\
& K-means-Only             & 35.00            &   3.71        &   43.00    &  5.04    \\
& Tinybenchmark            & 33.46            &  \textbf{2.17}         & 39.36      & \textbf{1.40}  \\
\bottomrule
\end{tabular}
}
\caption{Performance Comparison Tinybenchmark and Full Evaluation Benchmark. Delta is the absolute value of the difference between the baselines and the full Benchmark.}
\label{tab:estimate_error}
\end{table}
We compare the estimate error between three baseline methods and the real full benchmark. Random refer to randomly select samples from the full benchmark to form a tiny validation set. K-Means-Only refer to the simple average of the tinybenchmark $\mathcal{V}_{\text{tiny}}$ results. Tinybenchmark is the complete result we used in our main experiments which contains observed results and IRT predictions. We report the accuracy of the original Qwen2.5-7B. As shown in the table~\ref{tab:estimate_error}, tinybenchmark achieves the lowest delta value among all evaluation methods, proving that tinybenchmark can serve as a reliable proxy of full benchmark evaluation.
\section{Implementation Details}
\subsection{Implementation Details of Tinybenchmark}
The construction of Tinybenchmark requires extensive historical evaluation data from models to train the M-2PL model. For the MUSR and GPQA tasks, we crawled all available data from HuggingFace LLM OpenLeadboard~\footnote{{\url{https://huggingface.co/collections/open-llm-leaderboard/details}}}, and after cleaning, obtained results from a total of 432 models. For the GSM8K and ARC-C tasks, we collected data from HuggingFace LLM OpenLeadboard-Old~\footnote{{\url{https://huggingface.co/open-llm-leaderboard-old/datasets}}}, and compiled results from 395 models after cleaning. Using the Tinybenchmark code~\footnote{{\url{https://github.com/felipemaiapolo/tinyBenchmarks/tree/main}}}, we trained new M-2PL models for each task, ultimately resulting in our own tinybenchmark.
\subsection{Implementation Details of Main Experiment}
All the experiments are done under the zero-shot setting. We use the DeepSeek-V3.2-Chat API~\footnote{{\url{https://api.deepseek.com/v1}}} to synthetic reasoning step for both GSM8K and ARC-C dataset. We input the question and answer into DeepSeek and ask it to supplement the reasoning path.
\section{Prompts}
\begin{table}[H]
    \small
    \caption{Prompt template for In-Context Learning.}
    \centering
    \renewcommand{\arraystretch}{1.2}
    \begin{tabularx}{\linewidth}{>{\raggedright\arraybackslash}X}
        \toprule
        \# Instruction \\
        Below is a list of conversations between a human and an AI assistant (you). \\
As an AI assistant, you will engage in conversations with users, responding to their queries which are presented under the heading ``Query:''. \\
Your responses should be entered under the heading ``Answer:''. \\
You excel in a wide range of tasks including, but not limited to, providing general information, conducting reasoning, engaging in role-play, creative writing, planning, and solving mathematical and coding problems. \\
Your responses should be well-structured, comprehensive, and aim to thoroughly address the user's query or problem at hand. \\
(Demonstrations) \\
\# Query: \\
\# Answer: \\
...\\
(Question)\\
\# Query:\\
        \bottomrule
    \end{tabularx}
    \label{table_icl_prompt}
\end{table}
We use the prompt template from~\cite{zhaocontext} for In-Context Learning.
\begin{table}[H]
    \small
    \caption{Prompt template for reasoning path synthesis.}
    \centering
    \renewcommand{\arraystretch}{1.2}
    \begin{tabularx}{\linewidth}{>{\raggedright\arraybackslash}X}
        \toprule
\# Role \\
You are a reasoning path completion expert. Your task is to analyze the given "Question" and "Answer," then deduce and fill in the complete logical reasoning process that connects them.\\
\# Task\\
Based on the provided [Question] and [Answer], complete the missing [Reasoning Path] between them.\\
\# Output Requirements\\
Reasoning Path: Clearly and coherently present the step-by-step thought process, ensuring logical rigor.\\
Language: Use English.\\
Format: Begin with "Reasoning Path:" and present the steps using bullet points or a natural paragraph format.\\
\# Input Format\\
{}[Question]: [The user's question]\\
{}[Answer]: [The final answer to the question]\\
        \bottomrule
    \end{tabularx}
    \label{table_icl_prompt}
\end{table}
We use this template for reasoning path synthesis.
\end{document}